\documentclass[10pt,twocolumn,letterpaper]{article}

\usepackage[applications]{wacv}    

\usepackage{array}
\usepackage{booktabs}
\newcolumntype{L}[1]{>{\raggedright\arraybackslash}p{#1}}

\definecolor{wacvblue}{rgb}{0.21,0.49,0.74}
\usepackage[pagebackref,breaklinks,colorlinks,allcolors=wacvblue]{hyperref}

\def\wacvPaperID{3693} 
\def\confName{WACV}
\def\confYear{2027}

\title{Ego-OSCAR: Egocentric Open source Stereo CAptuRe System}

\author{
  Senthil Palanisamy \quad Satpal Singh Rathore \quad Pratyush Kumar Patnaik \\
  Shubhanshu Khatana \quad Gunjan Paul \quad Abhishek Anand \\
  FPV Labs \\
  \texttt{abhishek@fpvlabs.ai} \\[3pt]
  {\normalfont\normalsize All authors contributed equally.}
}

\begin{document}
\maketitle

\begin{abstract}
Egocentric vision has scaled its corpora far faster than its instruments. Calibrated stereo rigidly coupled to a synchronized inertial stream is not new; research-grade headsets deliver it. What does not exist is an open version: those platforms are closed products a third party cannot build or extend, and the consumer cameras used elsewhere are monocular and rolling-shutter. The instrument is the least reproducible part of an otherwise open pipeline. We present our system, an open-hardware head-mounted capture platform, at a complete bill of materials of $\sim$USD~200 in commodity parts and 3D prints. It pairs a hardware-synchronized global-shutter stereo camera with a 6-axis IMU, an embedded Linux SBC for on-device encoding, and a real-time microcontroller that bridges the camera and inertial clock domains to a residual of 700~\textmu s and watchdogs the recording against silent failure. We release the full stack alongside the device: CAD, firmware, recording and time-synchronization tooling, and our corpus, roughly 550 hours of stereo video over 1,462 sessions and 25 contributors, every device shipping its own stereo calibration. The corpus ships with model-predicted annotation layers rather than raw video alone, including per-frame 3D hand reconstructions spanning essentially the whole timeline; these are automatic predictions, not human-verified ground truth, and we report them as such. We characterize the instrument along sensor fidelity, data utility, and six months of distributed field deployment. Our system is not the highest-fidelity egocentric rig; it is the cheapest one a reader of this paper can reproduce, and the only corpus here that others can extend rather than only consume.
\end{abstract}

\section{Introduction}
\label{sec:intro}

\begin{figure*}[t]
    \centering
    \begin{subfigure}[b]{0.30\textwidth}
        \centering
        \includegraphics[width=\textwidth]{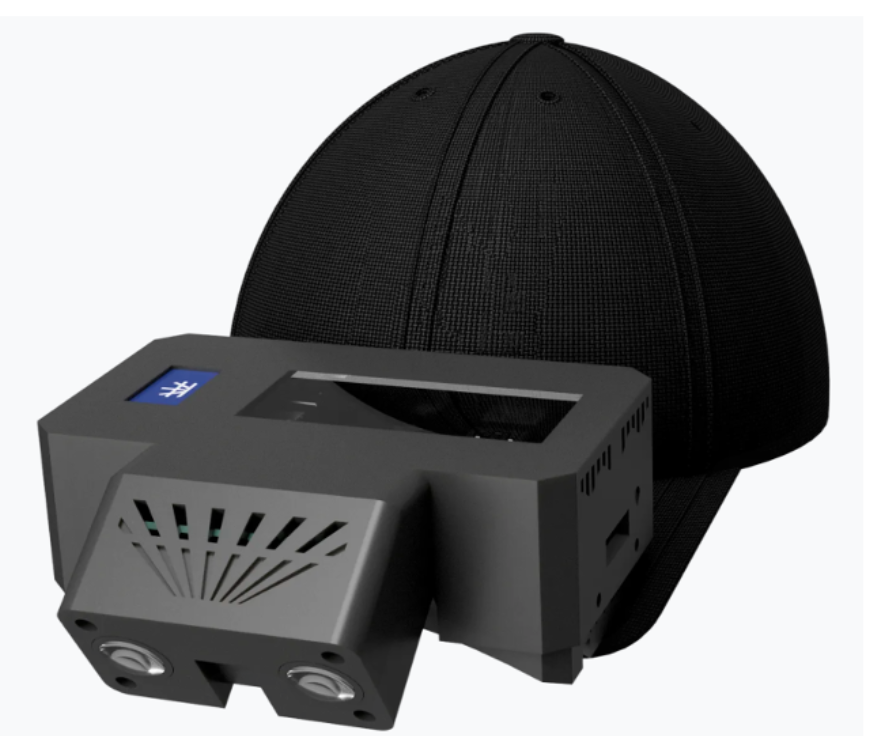}
        \caption{Assembled device}
        \label{fig:sub1}
    \end{subfigure}
    \hfill
    \begin{subfigure}[b]{0.40\textwidth}
        \centering
        \includegraphics[width=\textwidth]{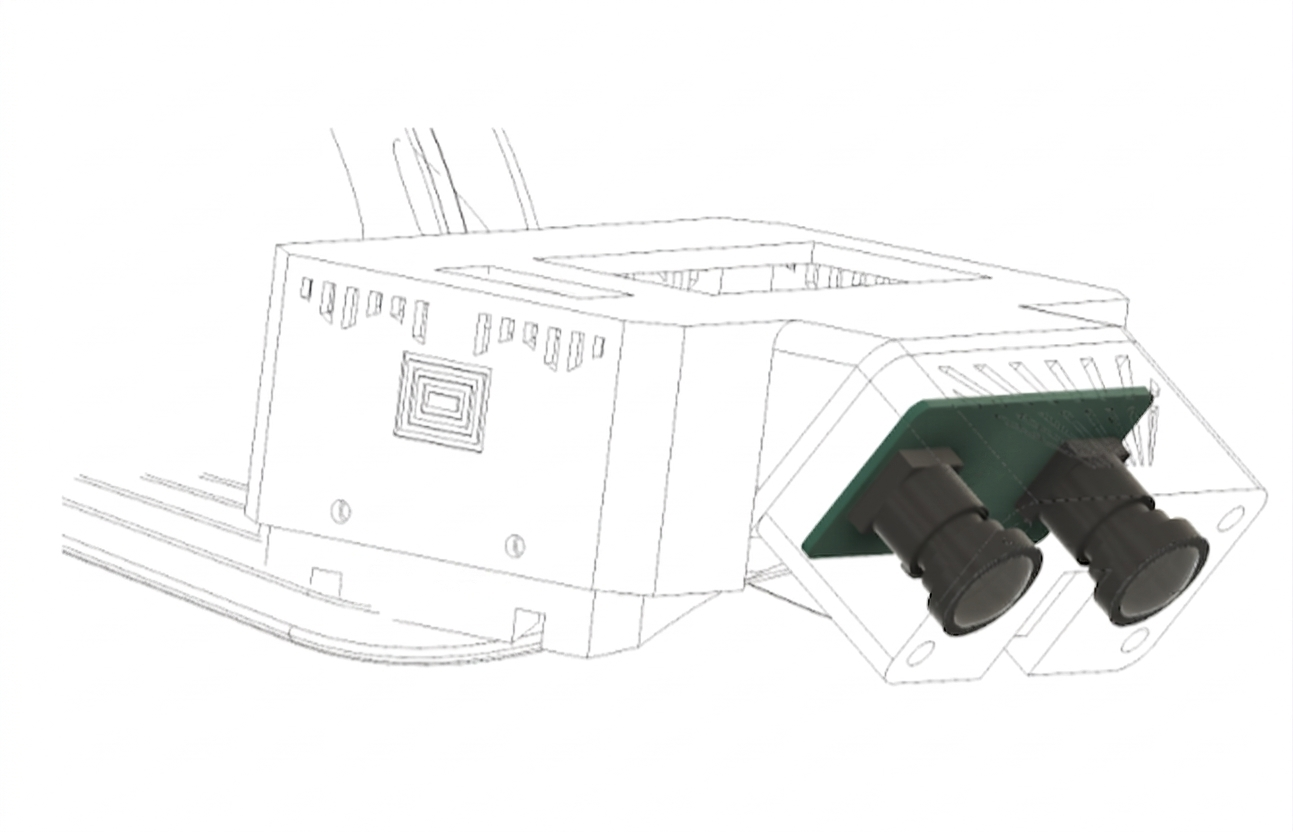}
        \caption{Mechanical outline}
        \label{fig:sub_mech}
    \end{subfigure}
    \hfill
    \begin{subfigure}[b]{0.20\textwidth}
        \centering
        \includegraphics[width=\textwidth]{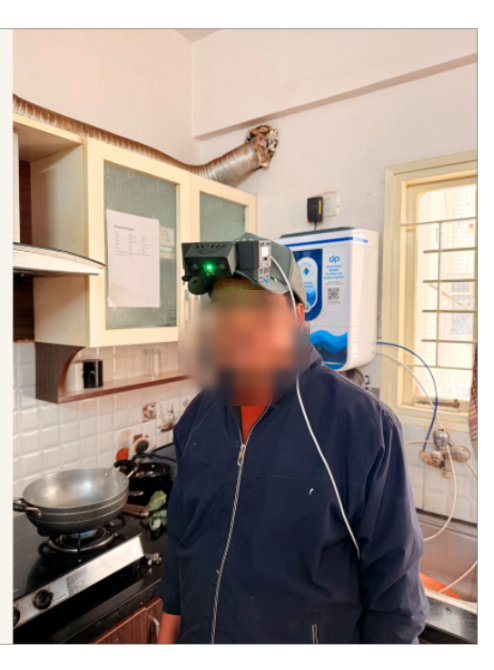}
        \caption{Worn in the field}
        \label{fig:sub2}
    \end{subfigure}
    \caption{\textbf{Our system.} A head-mounted, hardware-synchronized global-shutter stereo camera with a 6-axis IMU, an embedded SBC for on-device encoding, and a real-time microcontroller for clock bridging and watchdog. Total assembled mass is $\sim$280~g and the complete bill of materials is $\sim$USD~200 in commodity parts and 3D prints.}
    \label{fig:hero_image}
\end{figure*}

Egocentric vision has scaled its corpora much faster than it has scaled its instruments. Ego4D~\cite{Ego4D2022CVPR} established the modern scale of first-person data, Ego-Exo4D~\cite{EGOx4D} paired it with synchronized third-person views, and EPIC-KITCHENS~\cite{EPIC-KITCHEN} demonstrated the value of deep single-domain coverage. Across this progression the capture device has largely been treated as a fixed given rather than as a variable the community can design against, and the resulting hardware monoculture bounds what the data can support.

Two constraints follow from that monoculture. The first is geometric. Ego4D and EPIC-KITCHENS are monocular and rolling-shutter, with no calibration shipped alongside the video; rolling shutter in particular introduces per-row temporal offsets that corrupt geometry precisely under the rapid head rotation that characterizes egocentric capture~\cite{schubert2019rollingshutter}. Project Aria~\cite{ARIA}, which underpins Aria Everyday Activities~\cite{lv2024aria} and Nymeria~\cite{nymeria}, is a substantially better instrument, but its stereo pair is \textit{monochrome} and dedicated to SLAM: excellent for tracking, while the RGB stream that carries appearance remains a single view. For a large class of first-person problems, this is the binding constraint. Metric-scale 3D hand reconstruction, hand--object interaction, contact and grasp reasoning, and egocentric depth all want a scale that monocular RGB does not determine~\cite{zimmermann2017hand3d}; a calibrated RGB stereo pair supplies it directly, at the sensor, without a learned scale prior or a scene assumption.

The second constraint is reproducibility. Every corpus named above was captured on hardware that a reader of this paper cannot build or modify. Aria is distributed through a research access program; the consumer rigs used elsewhere are closed products whose shutter, exposure, and synchronization behavior are not under the experimenter's control. A dataset collected this way can be consumed but not extended: a group that wants a different environment distribution, a different activity mix, or a different sensor configuration has no path to producing compatible data. Hardware is thus the part of the pipeline contributing least to the progress that egocentric data collection needs.

We argue that the capture device deserves the same open-hardware treatment that ALOHA~\cite{zhao2023aloha} and UMI~\cite{UMI} brought to manipulation, and we present \textbf{our system} (Fig.~\ref{fig:hero_image}) as a concrete attempt: a head-mounted, hardware-synchronized global-shutter RGB stereo camera coupled to a 6-axis IMU, an embedded Linux SBC for on-device hardware-accelerated encoding, and a real-time microcontroller that bridges the camera and inertial clock domains and watchdogs the recording. The complete bill of materials is $\sim$USD~200, using only commercially available components and 3D-printed parts, with no custom PCB or proprietary silicon. The design deliberately declines on-device SLAM: pose estimation is treated as an offline batch problem so the hardware can optimize strictly for reliable ingestion on a wearable power budget.

A capture device has to be deployed before it can be assessed. We therefore also release \textbf{our corpus}, roughly 550 hours of stereo video, every hour of it captured simultaneously by both cameras, over 1{,}462 sessions captured by 25 contributors across 40+ indoor environments in a six-month field deployment, every session shipping the stereo calibration of the device that recorded it. The corpus ships with a per-frame 3D hand-reconstruction layer spanning essentially the entire timeline rather than raw video alone. This layer is model output, not human annotation, and we frame it throughout as a prediction rather than as ground truth.

We make three contributions:
\begin{itemize}
\itemsep0.15em
\item \textbf{An open-hardware egocentric capture device} delivering calibrated RGB stereo with a hardware-synchronized 6-axis IMU at a $\sim$USD~200 bill of materials, released with full CAD, wiring, and assembly documentation.
\item \textbf{An open capture stack}: hardware-accelerated recording daemon, IMU sampler, cross-clock-domain time-synchronization tooling, and watchdog firmware, characterized end-to-end in the field at a 96\% usable-session rate, with the residual visual--inertial offset measured in Sec.~\ref{sec:eval}.
\item \textbf{Our corpus}, 550 hours of stereo video with a corpus-wide predicted 3D hand-reconstruction layer, released with per-device calibration so that the geometry is usable rather than merely present.
\end{itemize}

We are explicit about scope. Whether egocentric human video transfers to downstream embodied and video-language tasks is a question prior work has already answered at scale~\cite{nair2022r3m, majumdar2023vc1, lin2022egovlp, shi2026egohumanoid}; what remains unsolved is how to \textit{capture} such video, with calibrated geometry and time-aligned inertial data, at a cost that permits deployment across hundreds of contributors rather than a single lab. This paper addresses the second question, and evaluates itself accordingly, as an instrument rather than as a method. We therefore do not claim parity with research-grade hardware on sensor fidelity, and we report the operating envelope of the device, including where off-the-shelf visual-inertial odometry fails on our data, rather than only where it succeeds (Sec.~\ref{sec:eval}). Nor do we claim that a model trained on this corpus specifically beats one trained on an existing corpus.

\section{Related Work}
\label{sec:related}

\paragraph{Egocentric datasets.}
Ego4D~\cite{Ego4D2022CVPR} established the modern scale of egocentric data with thousands of hours of unscripted activity across diverse environments, and Ego-Exo4D~\cite{EGOx4D} paired egocentric with exocentric views to enable skill-transfer research. EPIC-KITCHENS~\cite{EPIC-KITCHEN} preceded both with a deep single-domain kitchen collection annotated against a closed taxonomy of 97 verbs and 300 nouns. Aria Everyday Activities~\cite{lv2024aria} and Nymeria~\cite{nymeria} pair egocentric video with motion capture or 3D body pose for action grounding. These corpora are foundational, and we make no claim to compete with them on raw volume or wearer count. They are also expensive, collected with research-grade or closed hardware under programs that are not directly extensible by others, which is the specific gap this work addresses.

\paragraph{Egocentric capture hardware.}
The reference research-grade device is Project Aria~\cite{ARIA}: glasses carrying dual monochrome SLAM cameras, an RGB camera, eye tracking, dual IMUs, and microphones, distributed through Meta's research access program. Aria is high quality but closed, and cannot be freely reproduced or adapted. HoloLens and Magic Leap offer comparable sensor suites at substantially higher cost. Among consumer devices, GoPro has been the de-facto egocentric rig in much prior work including UMI~\cite{UMI}, but is monocular, rolling-shutter, and provides no hardware-synchronized inertial stream. EgoCap~\cite{Rhodin_EgoCap_2016} demonstrated head-mounted stereo egocentric capture with two fisheye cameras much earlier, but targeted marker-less body motion capture with a bespoke, non-reproducible rig rather than scalable collection. To our knowledge no open-source head-mounted device targeting calibrated stereo-inertial egocentric capture currently exists.

\paragraph{Geometry from egocentric video.}
The downstream methods our device is meant to serve are largely geometric. Stereo matching~\cite{hirschmuller2008sgbm, lipson2021raftstereo} recovers dense disparity given a calibrated, rectified pair; visual-inertial odometry~\cite{qin2019vinsfusion, campos2021orbslam3} recovers metric camera trajectory given time-aligned inertial measurements; and hand reconstruction methods such as WiLoR~\cite{potamias2024wilor} recover 3D hand pose that a calibrated baseline can triangulate rather than leave up to scale. Each of these places a requirement on the instrument: known intrinsics and extrinsics~\cite{zhang2000calibration}, global shutter~\cite{schubert2019rollingshutter}, and a bounded camera--IMU temporal offset~\cite{furgale2013kalibr}. Existing egocentric corpora satisfy these unevenly, and typically ship no calibration at all, which is what pushes users toward nominal factory intrinsics.

\paragraph{Downstream value of egocentric video.}
That first-person human video is a useful pretraining substrate is well established, and we build on that evidence rather than reproduce it. R3M~\cite{nair2022r3m} pretrains a visual representation on Ego4D and improves downstream manipulation success by over 20\% against training from scratch and over 10\% against CLIP and MoCo features, across 12 simulated tasks and a real arm learning from 20 demonstrations. VC-1~\cite{majumdar2023vc1} scales the recipe to over 4{,}000 hours of egocentric video drawn from seven sources and evaluates it across the 17 tasks of CortexBench. EgoVLP~\cite{lin2022egovlp} establishes the same for egocentric video-language pretraining, transferring to five downstream tasks across three egocentric benchmarks. Most directly, EgoHumanoid~\cite{shi2026egohumanoid} co-trains a vision-language-action policy on robot-free egocentric human demonstrations with a small amount of robot data and reports a 51\% improvement over robot-only baselines in unseen environments. Given this evidence, a further demonstration that egocentric video transfers would add little; making such video cheaper to capture would add more. This paper does the latter.

\paragraph{Lowering the capture barrier.}
MobileEgo Anywhere~\cite{palanisamy2026mobileego} attacks the same accessibility problem from the opposite direction, using commodity smartphone sensing for long-horizon capture and releasing 200 hours across 584 sessions. Smartphones remove the hardware barrier entirely, at the cost of no hardware-level stereo synchronization and no control over shutter or exposure timing. Our system sits one step up that cost curve, accepting a $\sim$USD~200 bill of materials for a hardware-synchronized global-shutter stereo pair and per-device calibration: complementary points on an accessibility--fidelity frontier.

\paragraph{Open-source hardware.}
Our system follows the lineage of open hardware that has reshaped robotics practice. ALOHA~\cite{zhao2023aloha} and Mobile ALOHA~\cite{fu2025mobile} released a complete bimanual teleoperation rig that labs could replicate directly; UMI~\cite{UMI} applied the same philosophy to handheld manipulation capture; DexCap~\cite{wang2024dexcap} to portable hand motion capture. We aim to occupy the analogous role for head-mounted, observation-only egocentric capture. In each case the released hardware, as much as the method, is what let other groups build on the work.

\section{System Design}
\label{sec:method}

Our system comprises a head-mounted capture device (Fig.~\ref{fig:hero_image}) and an accompanying capture pipeline. The system is engineered around four pillars: \textbf{affordability} (COTS-only bill of materials, no custom PCBs or proprietary silicon); \textbf{reproducibility} (open schematics and a 3D-printable enclosure); \textbf{sensor fidelity} (hardware-synchronized stereo video and high-rate inertial data for visual-inertial estimation); and \textbf{deployment robustness} (real-time fault detection to maximize data yield across a distributed contributor network). Rather than run power-hungry on-device SLAM, we treat pose estimation as an offline batch problem, which lets the hardware optimize strictly for reliable ingestion.

\subsection{Hardware}
\label{sec:hardware}

\begin{figure}[t]
    \centering
    \includegraphics[width=\linewidth]{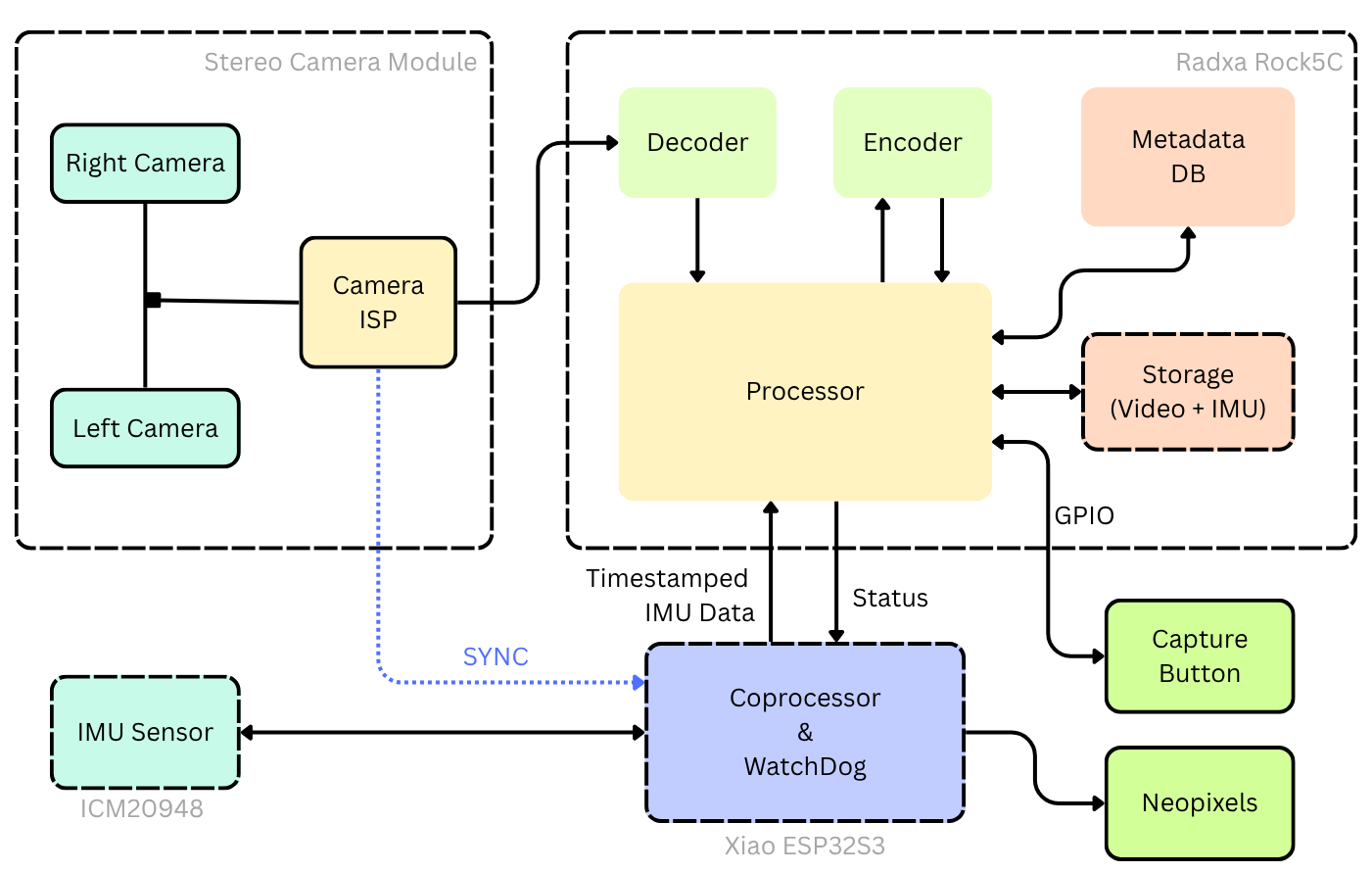}
    \caption{\textbf{Hardware architecture.} The stereo camera's Start-of-Exposure pin drives an ESP32 interrupt, placing frame timing and IMU sampling in a single clock domain that is later aligned to video frame indices offline.}
    \label{fig:hardware_design}
\end{figure}

The device (Fig.~\ref{fig:hardware_design}) is an all-in-one head-worn unit: a hardware-synchronized stereo camera, a single-board computer (SBC) for on-device encoding and storage, a 6-axis IMU, and a real-time microcontroller for timestamping, user interface, and watchdog duty. Total assembled mass is $\sim$280~g, mounted via a commodity sport visor. The requirements span three axes: the \textit{observation side} (sufficient visual context, stereo depth, robustness to rapid head motion); the \textit{measurement side} (time-aligned inertial data for visual-inertial fusion); and the \textit{operational side} (autonomous battery operation, real-time fault surfacing, field durability).

\paragraph{HD1. Hardware-synchronized global-shutter stereo camera.}
We use a Dexcin USB stereo module: two Omnivision global-shutter sensors hardware-synchronized through a single ASIC, exposing one USB~2.0 endpoint to the host. The sensors capture 1280$\times$720~px per camera at 30~FPS, delivered as a single side-by-side stitched MJPEG frame. Per-sensor FOV is 126\textdegree{} and the baseline is 42~mm. Four properties drove the choice. \textit{Global shutter} eliminates the per-row temporal offsets that corrupt visual-inertial estimation under rapid head motion~\cite{schubert2019rollingshutter}. \textit{Hardware sync} between left and right removes the microsecond-to-millisecond stereo offsets that break depth estimation on moving scenes. A \textit{single USB endpoint} over UVC ensures compatibility with stock Linux drivers without dual-cable routing. Finally, an \textit{SoE trigger output} exposes the Start-of-Exposure signal on a STRB pin, which we use to bridge the camera's internal clock domain to the microcontroller (CD3 below). Unlike monocular wide-FOV approaches that recover implicit stereo via side mirrors (e.g., UMI~\cite{UMI}), our system captures genuine binocular disparity at the sensor, so standard calibration and rectification pipelines apply unmodified.

\paragraph{HD2. Embedded SBC with hardware video acceleration.}
The recording host is a Radxa Rock~5C (Rockchip RK3588, 2~GB RAM), which provides hardware acceleration for both MJPEG decode and H.264 encode, sufficient to sustain 30~FPS stereo without frame drops on a wearable power budget. The tradeoff space is narrow: the Raspberry Pi~5 lacks hardware MJPEG decode, while NVIDIA Jetson Nano and Orin Nano offer stronger compute but draw substantially more power and are physically larger. The RK3588 sits at the inflection point where hardware media acceleration meets wearable power and form-factor constraints.

\paragraph{HD3. Consumer-grade IMU.}
The inertial sensor is a TDK InvenSense ICM-20948, from which we record 6 axes (3-axis accelerometer, 3-axis gyroscope) over I\textsuperscript{2}C at 180~Hz. The choice is deliberate: the part sits on a standard I\textsuperscript{2}C bus, so a higher-grade replacement (BMI088, ISM330) requires no firmware change beyond the driver. For the tasks we target, the IMU is a secondary signal that assists visual estimation in low-feature regimes and provides gravity alignment, rather than serving as the sole pose estimator. A higher-grade part adds USD~80--200 to the bill of materials, which is material when the fleet is measured in hundreds of units.

\paragraph{HD4. Real-time microcontroller.}
A Seeed Studio Xiao ESP32-S3 connects to the Radxa over UART and serves three roles. For \textit{clock bridging}, the camera module and IMU run on independent clocks with substantial observed offset and drift; the ESP32 taps the camera's SoE signal into an interrupt handler that records a monotonic timestamp on every exposure, reads the IMU in the same domain, and forwards the merged stream to the SBC over UART. For \textit{user feedback}, the MCU drives an RGB LED strip signalling device state (booting, ready, recording, error) from a 1~Hz heartbeat. For \textit{watchdog} duty, absence of that heartbeat for more than two seconds raises a visible error state. Without this last function a wearer can record for an hour after the SBC has hung without noticing, a failure mode we observed repeatedly in early deployments.

\paragraph{HD5. Power, enclosure, and durability.}
The device is powered over USB-PD from a 10{,}000~mAh power bank giving 5--6~hours of sustained capture, hot-swappable for all-day sessions. The enclosure is three 3D-printed parts. Field deployment surfaced three durability issues: pressure points along the visor strap during prolonged wear; a forward center of mass that causes the visor to droop over multi-hour sessions; and no moisture resistance in the current enclosure. The complete bill of materials, at \textbf{$\sim$USD~200 (INR~19{,}600)}, is given in Tab.~\ref{tab:bom}.

\begin{table}[t]
\centering
\caption{Bill of materials. All parts are commercially available; no custom PCB or proprietary silicon is required.}
\label{tab:bom}
\small
\setlength{\tabcolsep}{4pt}
\begin{tabular}{@{}lrl@{}}
\toprule
\textbf{Component} & \textbf{INR} & \textbf{Function} \\
\midrule
Dexcin USB stereo camera & 6{,}300 & Global-shutter stereo \\
Radxa Rock 5C (2~GB) & 6{,}500 & Capture, encode, store \\
Heatsink (Radxa) & 800 & Thermal management \\
256~GB SD card & 2{,}500 & OS $+$ $\sim$18~h recording \\
USB cable, A--C 90\textdegree & 300 & Camera $\leftrightarrow$ SBC \\
ICM-20948 6-axis IMU & 800 & Inertial sensing \\
Seeed Xiao ESP32-S3 & 650 & Sync, UX, watchdog \\
Misc.\ electronics & 300 & LEDs, buzzer, wiring \\
3D-printed shells, screws & 300 & Enclosure \\
Visor / cap & 300 & Head mount \\
USB-C PD cable, 90\textdegree & 150 & SBC $\leftrightarrow$ power \\
10{,}000~mAh power bank & 1{,}000 & Power \\
\midrule
\textbf{Total} & \textbf{19{,}600} & \textbf{$\sim$USD~205} \\
\bottomrule
\end{tabular}
\end{table}

\subsection{Capture Pipeline}
\label{sec:pipeline}

\paragraph{CD1. Hardware-accelerated video.}
Recording uses a modified ffmpeg build that invokes Rockchip's Media Process Platform for hardware MJPEG decode and H.264 encode. Output is MP4 at 12--14~GB/hour, roughly a 5$\times$ reduction from raw MJPEG (70--80~GB/hour) with no perceptible quality loss. The hardware path is essential: software encoding on the RK3588 cannot sustain 30~FPS stereo without dropping frames or thermally throttling.

\paragraph{CD2. Clip segmentation.}
Recordings are segmented into 5-minute MP4 clips, which bounds data loss on power failure to one trailing clip per session and enables incremental processing of long sessions.

\paragraph{CD3. Time synchronization.}
The camera and IMU run on independent clocks, with substantial offset and drift observed in testing. The ESP32 bridges the gap as described in HD4, and an offline pass aligns its frame timestamps to video frame indices. Because the camera pipeline occasionally drops or duplicates frames at startup, a one-to-one interrupt-to-frame mapping is not guaranteed. To anchor both streams against a known reference, a blue LED mounted beside the camera lens is wired to the ESP32 and flashes on exactly the 60th interrupt; detecting that flash in the video pins the 60th interrupt to its true frame index, after which the remainder of the sequence aligns deterministically. We characterize the residual offset that survives this correction in Sec.~\ref{sec:eval}.

\paragraph{CD4. Watchdog and storage.}
The 1~Hz UART heartbeat doubles as the watchdog signal; loss for more than two seconds triggers the error state, empirically catching pipeline hangs from USB driver edge cases, thermal throttling that escalates to a system freeze, and SD card I/O errors. Recordings are written to a 256~GB SD card ($\sim$18~hours capacity), and a post-session daemon handles upload to NAS, S3, or GCS with resume support and integrity verification.

\section{Characterizing the Instrument}
\label{sec:eval}

Because the contribution is a capture system rather than a method, we evaluate it as an instrument, along three tiers matching its design goals: \textbf{sensor fidelity}, whether the raw streams meet the requirements that downstream egocentric pipelines place on the hardware; \textbf{data utility}, whether the captured data supports representative first-person tasks; and \textbf{deployment scale}, whether the system is operationally reliable when distributed across a real contributor network.

\begin{figure*}[t]
    \centering
    \begin{subfigure}[b]{0.30\textwidth}
        \centering
        \includegraphics[width=\textwidth]{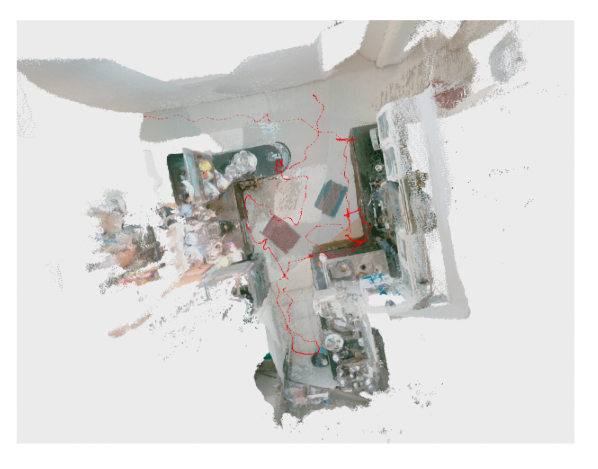}
        \caption{VINS-Fusion~\cite{qin2019vinsfusion} trajectory overlaid on a Kinect point cloud}
        \label{fig:vins_trajectory}
    \end{subfigure}
    \hfill
    \begin{subfigure}[b]{0.66\textwidth}
        \centering
        \includegraphics[width=\textwidth]{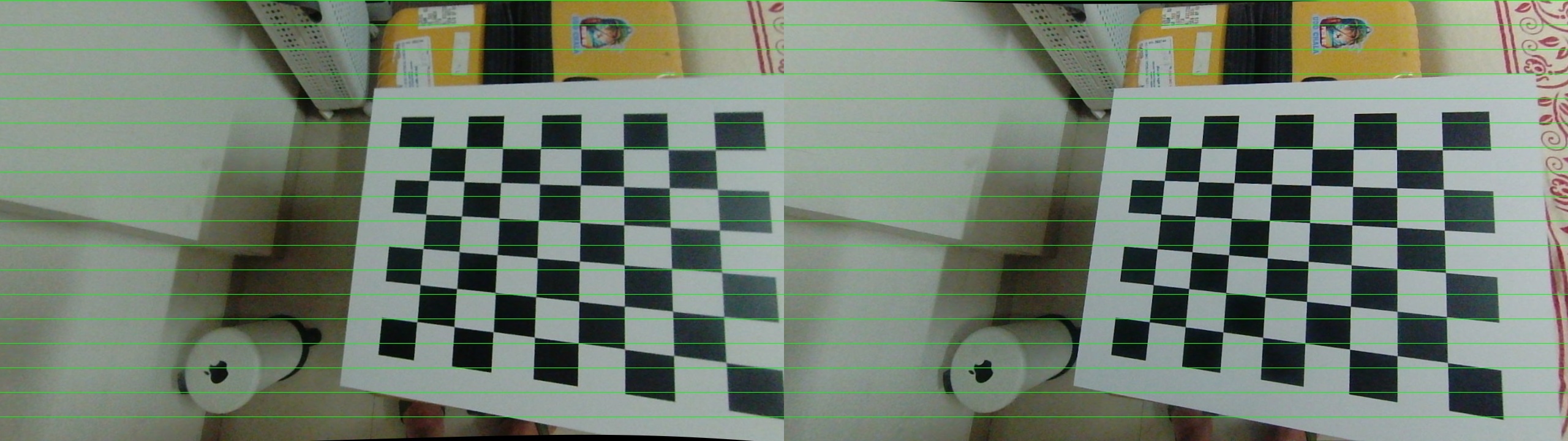}
        \caption{Rectified stereo pair, epipolar lines overlaid}
        \label{fig:rectified}
    \end{subfigure}
    \caption{\textbf{Instrument characterization.} (a) Off-the-shelf visual-inertial odometry on a capture from our system. The trajectory is drawn over a Kinect point cloud for visualization only: the two were not registered or scaled to each other, so the overlay is qualitative and asserts no quantitative alignment. We report this run as a convergence rate rather than a trajectory-accuracy result, since we hold no surveyed ground truth for these sequences. (b) After calibration, corresponding scene points fall on common scanlines across the pair, the precondition for standard stereo matching; the mean per-pixel epipolar error across the fleet is 0.4~px.}
    \label{fig:depth}
\end{figure*}

\subsection{Tier 1: Sensor Fidelity}

\paragraph{Stereo geometry.}
Calibration is performed per device rather than per device model, which matters for a fleet of hand-assembled units whose optical alignment differs from unit to unit. Each of the 13 devices is calibrated from an 8$\times$6 chessboard with 30~mm squares at the native 1280$\times$720 capture resolution, fitting a pinhole model with radial-tangential distortion (five coefficients $k_1$, $k_2$, $p_1$, $p_2$, $k_3$) per camera plus a $3\times3$ rotation, a translation, and the $\approx$42~mm baseline~\cite{zhang2000calibration}. Per-camera reprojection error is below 0.03~px, and each device's calibration ships as a \texttt{calibration.json} with every session it recorded, so that downstream users never fall back on nominal factory intrinsics. Across all 13 deployed devices the mean per-pixel epipolar error after rectification is 0.4~px (Fig.~\ref{fig:rectified}). The 126\textdegree{} per-sensor FOV introduces predictable barrel distortion at the periphery, which the calibration removes cleanly.

\paragraph{Inertial noise.}
We characterize the ICM-20948 with an Allan variance protocol~\cite{elsheimy2008allan} over 12 hours of stationary capture. Accelerometer noise density is $3.64 \times 10^{-2}$~m/s\textsuperscript{2}/$\sqrt{\text{Hz}}$ and gyroscope bias instability is $9.68 \times 10^{-4}$~rad/s, placing the part in the consumer-grade range and comparable to phone-grade IMUs. For gravity alignment, motion classification, and visual-inertial fusion as a coarse prior, this noise floor is acceptable; for long-horizon inertial integration we recommend the higher-grade swap described in Sec.~\ref{sec:hardware}.

\paragraph{Visual--inertial synchronization.}
After the per-session offset correction of Sec.~\ref{sec:pipeline}, the residual lag between the visual and inertial streams is \textbf{700~\textmu s}, measured with the Kalibr~\cite{furgale2013kalibr} camera--IMU offset test.

\subsection{Tier 2: Data Utility}

\paragraph{Stereo depth.}
We run SGBM~\cite{hirschmuller2008sgbm} and RAFT-Stereo~\cite{lipson2021raftstereo} on held-out sequences and recover dense disparity across the full FOV without specialized tuning, both methods consuming the released calibration directly. The 42~mm baseline and 126\textdegree{} FOV give reliable depth over the 0.5--4~m range typical of indoor egocentric capture. We report this qualitatively: absent depth ground truth for unscripted household capture, we make no claim about metric depth accuracy, only that the rectified pair supports standard matching pipelines unmodified.

\paragraph{Visual odometry.}
We run VINS-Fusion in stereo-inertial mode~\cite{qin2019vinsfusion} on 20 held-out sequences of 1--3 minutes: 12/20 produce stable trajectories (Fig.~\ref{fig:vins_trajectory}) while 8 diverge, from a combination of brittleness under the dynamic scene content and rapid head motion characteristic of egocentric capture, and inertial bias accumulation consistent with the noise floor above. On the same environments and activities an Intel RealSense reaches 15/20, holding an advantage by virtue of active stereo depth.

We are deliberate about what this number is. It is a \textbf{convergence rate}, not a trajectory-accuracy result: we have no motion capture or surveyed ground truth for these sequences, so we report no ATE or RPE, and a reader should not infer metric pose quality from 12/20. It characterizes the operating envelope of the current device, showing where off-the-shelf visual-inertial odometry does and does not hold up on this data, which is also why we do not ship a camera-trajectory annotation layer. A ground-truthed evaluation requires an instrumented capture campaign we have not yet run, and we list it as a limitation rather than approximating it here.

\paragraph{Hand detection.}
We run WiLoR~\cite{potamias2024wilor} hand detection and 3D reconstruction over the full corpus. We define the \textbf{hand-detection rate} precisely, as the metric is otherwise ambiguous: the fraction of decoded video frames in which WiLoR returns at least one hand instance above its default confidence threshold, computed over the whole corpus rather than a sample. By that definition the rate is \textbf{94\%}. This is a \textit{coverage} statistic, not an accuracy one: we hold no manually annotated hand ground truth here, so we make no claim about the geometric fidelity of the reconstructions. It supports one narrow conclusion, that the 126\textdegree{} FOV and head-mounted geometry keep near-field hand--object interaction in frame for the large majority of recorded time, which is a precondition for downstream hand-pose and manipulation work rather than a demonstration of it.

\begin{table*}[!t]
\centering
\caption{Comparison with existing egocentric corpora. Figures are as reported by each dataset's own publication and count unique content hours rather than per-sensor hours, so our 550 is the recorded wall-clock duration and not a doubling across the stereo pair; Ego-Exo4D's total combines egocentric and exocentric video. We are not competitive on raw volume or wearer count; the differences that matter are in sensing and in reproducibility of the instrument.}
\label{tab:comparison}
\footnotesize
\setlength{\tabcolsep}{3pt}
\begin{tabular}{@{}L{2.4cm}L{1.3cm}L{1.5cm}L{6.2cm}L{1.6cm}L{2.3cm}@{}}
\toprule
\textbf{Dataset} & \textbf{Hours} & \textbf{Wearers} & \textbf{Camera} & \textbf{Sync.\ IMU} & \textbf{Open HW} \\
\midrule
Ego4D~\cite{Ego4D2022CVPR} & 3{,}670 & 931 & Mono consumer, rolling shutter (stereo in a subset) & Partial & No \\
\addlinespace
Ego-Exo4D~\cite{EGOx4D} & 1{,}286 & 740 & Aria: mono RGB $+$ 2 mono SLAM & Yes & No \\
\addlinespace
EPIC-K.-100~\cite{EPIC-KITCHEN} & 100 & 37 & Mono head-mounted & No & No \\
\addlinespace
Nymeria~\cite{nymeria} & 300 & 264 & Aria $+$ body mocap & Yes & No \\
\addlinespace
\textbf{Ours} & \textbf{550} & \textbf{25} & \textbf{Calibrated RGB stereo, global shutter, per-device calibration} & \textbf{Yes}, 180~Hz & \textbf{Yes}, $\sim$USD~200 \\
\bottomrule
\end{tabular}
\end{table*}

\subsection{Quality Control}
\label{sec:qc}

Data reaching the release passes three successive filters, which we state explicitly because the rejection criteria determine what a distributed-collection corpus actually contains. \textit{At capture}, the watchdog (Sec.~\ref{sec:pipeline}) catches a session that has silently stopped recording: loss of the heartbeat for more than two seconds puts the device into a visible error state, so the wearer stops rather than continuing for an hour against a hung SBC. \textit{Per batch}, every upload is validated for decodability, expected clip count and duration, and presence of the companion inertial and calibration artifacts. \textit{At selection}, sessions entering the release are screened for hand visibility, which is what makes the corpus consistently rich in hand--object interaction rather than merely long; sessions failing calibration or lacking a usable synchronized trace are excluded rather than shipped with caveats. The IMU is present in 1{,}271 of the 1{,}462 released sessions (86.9\%), and the remainder are released without an inertial stream rather than with an unverified one. The end-to-end effect of these filters is the 96\% usable-session rate below.

\subsection{Tier 3: Deployment Scale}

Across all sessions in the six-month deployment, \textbf{96\%} produced usable data end-to-end. Three failure modes dominated: thermal shutdown during sessions exceeding 90~minutes above 35\textdegree C, resolved by adding the heatsink in Tab.~\ref{tab:bom}; SD card I/O errors, resolved by per-batch validation; and cable strain at the camera-to-SBC connector, resolved by mechanical reinforcement. The watchdog protocol detected all three classes in active deployment, preventing wearers from unknowingly continuing a session with nothing being saved. The resulting yield, 550 hours of recorded content across 1{,}462 sessions, 40+ environments, and 25 contributors, is the primary evidence that our system works as a distributed, field-deployable capture platform rather than only under lab conditions.

\section{Our Corpus}
\label{sec:dataset}

To validate our system at deployment scale we ran it across a contributor network in India over roughly six months, collecting \textbf{1{,}462 stereo sessions} totalling approximately \textbf{550 hours of egocentric video}, every hour of it captured simultaneously by both cameras, across 40+ indoor environments, predominantly residential. The corpus is kitchen-centric but not kitchen-only: roughly a third of recorded time falls outside cooking and dishwashing, in domains such as sewing and tailoring that are largely absent from existing egocentric corpora. Sessions range from short focused tasks to hour-plus continuous activity, with a median of 14.9~minutes and a longest session of 211~minutes, giving both clean short episodes and long-horizon sequences from one pipeline.

\subsection{Predicted Annotation Layers}
\label{sec:annotations}

The release ships with machine-generated layers over the video rather than raw footage alone. Every layer is a \textit{prediction}: none has been manually annotated or verified against human labels, and none should be read as ground truth.

We run WiLoR~\cite{potamias2024wilor} over the full corpus and release the per-frame output for the whole release rather than a curated subset, at the 94\% coverage rate defined in Sec.~\ref{sec:eval}. Because the pair is calibrated and rectified, reconstructions can be triangulated across the stereo baseline rather than relying on monocular scale. A second layer of automatically generated open-vocabulary action segments ships alongside it and is characterized in the supplementary material. Both are offered as a convenience over the released video and as a starting point others can correct or replace; neither carries any part of the instrument claim we evaluate here.

\subsection{Task Diversity}
\label{sec:diversity}

The action-segment layer yields \textbf{209{,}315 predicted segments} covering $\approx$100\% of the recorded timeline at a median of 94 per session, which is dense temporal coverage rather than clip-level tagging. The vocabulary is open rather than a fixed taxonomy: 460 action verbs and 32{,}630 object phrases combine into 57{,}104 distinct verb--object pairs, and segment ordering exposes 192{,}509 unique task transitions. The leading verbs, \textit{place}, \textit{pick}, \textit{wash}, \textit{adjust}, \textit{rinse}, \textit{stir}, \textit{scrub}, \textit{wipe}, \textit{pour}, \textit{add}, \textit{cut}, and \textit{carry}, cover contact-rich interaction, containment, state change, and deformable-material handling, while the distribution stays long-tailed: the top-20 expressions account for only 1.5\% of instances, ranks 21--500 add a further 6.5\%, and the remaining 92.0\% spreads across thousands of distinct expressions.

\subsection{Positioning}

Tab.~\ref{tab:comparison} places the release against existing egocentric corpora. On \textit{sensing}, Ego4D and EPIC-KITCHENS are monocular and rolling-shutter, while Ego-Exo4D and Nymeria inherit Aria's suite of an RGB camera plus two \textit{monochrome} SLAM cameras, excellent for tracking but not calibrated RGB stereo. On \textit{reproducibility}, every other row was captured on hardware a third party cannot build or extend; this is the only corpus in the table that others can extend rather than only consume.

\subsection{Ethics, Consent, and Privacy}
\label{sec:ethics}

All capture was performed by consenting contributors in spaces where the environment owner granted permission, and contributors recorded their own routine activity rather than bystanders in public space. That bounds the exposure without eliminating it: household members and visitors do appear. Every session therefore passes a face-detection-and-blur pass before release that also targets screens, the two most direct carriers of identity and private content, and contributor identity in the released metadata is a salted hash rather than a name or serial.

\section{Released Artifacts}
\label{sec:artifacts}

\begin{figure}[t]
    \centering
    \includegraphics[width=\linewidth]{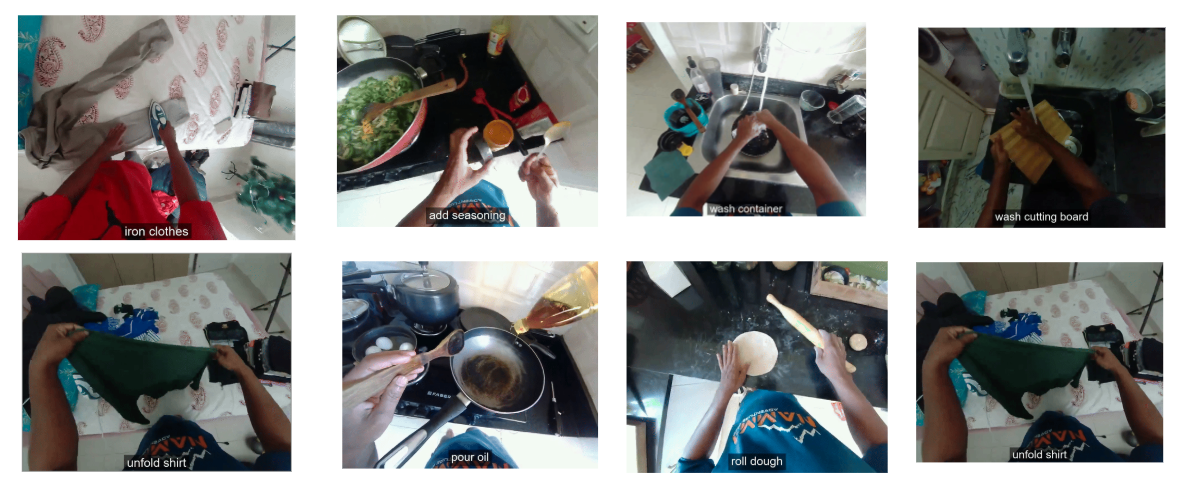}
    \caption{\textbf{Corpus samples.} Left-camera frames with their predicted action captions, drawn from the cooking, dishwashing, and laundry domains. Captions are free-form descriptive spans rather than entries in a fixed taxonomy, and the head-mounted geometry keeps hand--object interaction in frame.}
    \label{fig:corpus_samples}
\end{figure}

Everything behind the claims in this paper is released under a permissive license, at two entry points.

\vspace{0.3em}
\noindent\textbf{Capture device} --- \href{https://egocentric-hardwaree.vercel.app/stereo-cap}{egocentric-hardwaree.vercel.app/stereo-cap}. CAD sources and assembly guide, the bill of materials with supplier part numbers (Tab.~\ref{tab:bom}) and wiring diagram, the ESP32-S3 firmware (SoE interrupt handler, IMU sampling, UART framing, LED anchor, watchdog), the capture stack (recording daemon, hardware-accelerated encoder invocation, clip segmenter, uploader), and the calibration and time-synchronization tooling.

\vspace{0.3em}
\noindent\textbf{Corpus} --- \href{https://egocentric-hardware.vercel.app/dataset}{egocentric-hardware.vercel.app/dataset}. All 1{,}462 sessions of calibrated stereo video, each with its \texttt{calibration.json} and time-aligned inertial trace where present, plus the predicted annotation layers of Sec.~\ref{sec:annotations}. Fig.~\ref{fig:corpus_samples} shows representative frames with their predicted captions.

\vspace{0.3em}
\noindent The supplementary material gives the full corpus characterization: per-session contents, the most frequent action expressions, activity and environment composition, contributor diversity, long-tail structure, and calibration detail.

\section{Limitations}
\label{sec:limitations}

We validate the sensor package, not this corpus's value for representation learning: prior work establishes that egocentric human video transfers~\cite{nair2022r3m, majumdar2023vc1, lin2022egovlp, shi2026egohumanoid}, but that evidence concerns the modality rather than this corpus, and showing that a model trained on our data beats one trained on an existing corpus is the most important follow-on work. We hold no pose or hand ground truth, so we report visual-odometry convergence rather than ATE or RPE, and the 94\% hand figure is coverage rather than accuracy; both released annotation layers are model output, unscored against human labels, and should be treated as pseudo-labels. The corpus is also concentrated: 25 contributors in India across 13 shared devices, in a domestic rather than open-domain activity mix. On hardware, the consumer-grade IMU is the dominant pose-error source and is swappable on the same I\textsuperscript{2}C bus but not yet benchmarked in that configuration; the enclosure has the limitations noted in Sec.~\ref{sec:hardware}; and the device is capture-only, so a bad session cannot be rejected in real time.

\section{Conclusion}

We present a low-cost, open-hardware stereo-inertial capture system for egocentric vision at scale. At a complete bill of materials of $\sim$USD~200 it delivers hardware-synchronized global-shutter RGB stereo with per-device calibration and a time-aligned 6-axis inertial stream, from commodity components and 3D-printed parts. The accompanying capture stack bridges the camera and inertial clock domains and sustains a 96\% usable-session rate across a six-month distributed deployment. Our corpus, 1{,}462 sessions and $\approx$550 hours of stereo video from 25 contributors, shipped with a corpus-wide predicted 3D hand-reconstruction layer, validates the system at that scale. Our aim is to do for egocentric capture what open hardware has begun to do for manipulation: commoditize the instrument, so the field can spend its effort on the data and the models, not the rig.

{
    \small
    \bibliographystyle{ieeenat_fullname}
    \bibliography{main}
}

\end{document}